\documentclass[sigconf]{acmart}

\renewcommand\footnotetextcopyrightpermission[1]{} 

\usepackage{algorithm}
\usepackage{multirow}
\usepackage{caption}
\usepackage{subcaption}
\usepackage{booktabs}
\usepackage{algpseudocode}
\usepackage{tabularx}
\usepackage{xcolor,colortbl}
\usepackage{amsmath}

\begin{document}

\title{VidCoM: Fast Video Comprehension through Large Language Models with Multimodal Tools}

\author{Ji Qi\footnotemark[1], Kaixuan Ji\footnotemark[1], Jifan Yu, Duokang Wang, Bin Xu\footnotemark[2], Lei Hou, Juanzi Li}
\affiliation{
  Department of Computer Science and Technology, Tsinghua University
  \country{Beijing 100084, China}
  }
\email{
    {qj20, jkx19}@mails.tsinghua.edu.cn
}

\renewcommand{\shortauthors}{Ji Qi et al.}

\begin{abstract}
Building models that comprehends videos and responds specific user instructions is a practical and challenging topic, as it requires mastery of both vision understanding and knowledge reasoning.
Compared to language and image modalities, training efficiency remains a serious problem as existing studies train models on massive sparse videos paired with brief descriptions.
In this paper, we introduce \textbf{VidCoM}, a fast adaptive framework that leverages Large Language Models (LLMs) to reason about videos using lightweight visual tools.
Specifically, we reveal that the key to responding to specific instructions is focusing on relevant video events, and utilize two visual tools, structured scene graph generation and descriptive image caption generation, to gather and represent the event information.
Thus, a LLM enriched with world knowledge is adopted as the reasoning agent to achieve the responses by performing multiple reasoning steps on specific video events.
To address the difficulty of LLMs identifying video events, we further propose an Instruction-oriented Video Events Recognition (InsOVER) algorithm. This algorithm locates the corresponding video events based on an efficient Hungarian matching between decompositions of linguistic instructions and video events, thereby enabling LLMs to interact effectively with extended videos.
Extensive experiments on two typical video comprehension tasks show that the proposed tuning-free framework outperforms the pre-trained models including Flamingo-80B, to achieve the state-of-the-art performance.
Our source code and system will be publicly available.
\end{abstract}



\begin{teaserfigure}
  \includegraphics[width=\textwidth]{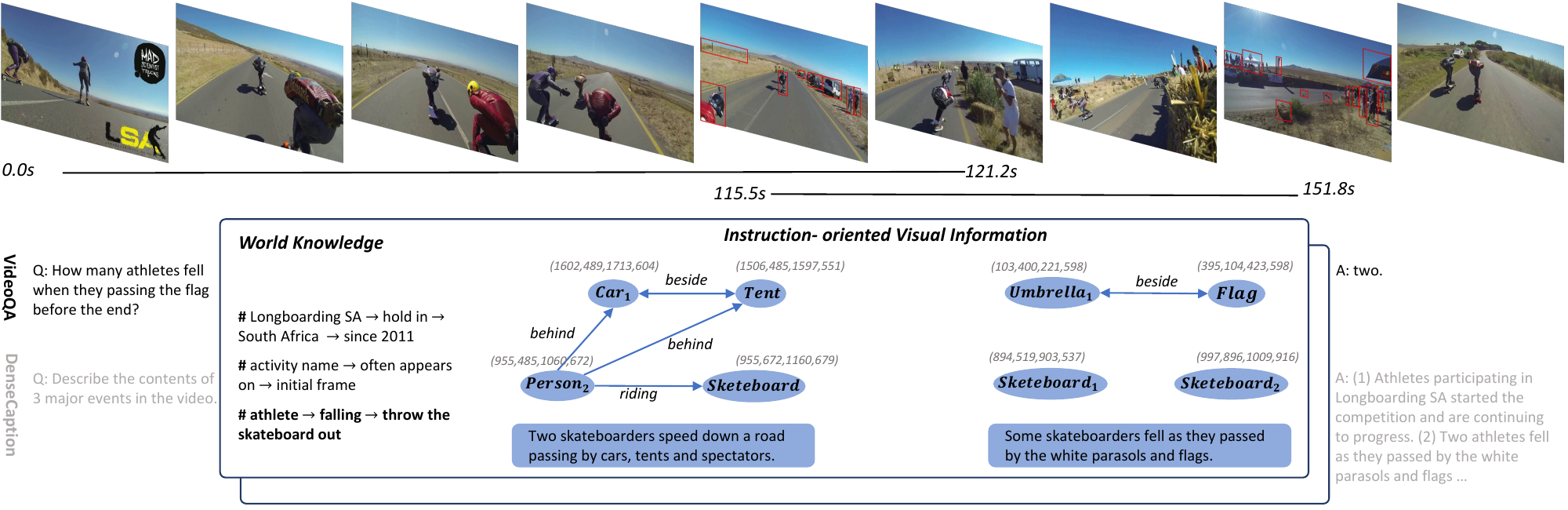}
  \caption{Given a specific user instruction, BiLL-VTG adopt a LLM agent to perform multiple reasoning steps on the video, where each step the agent acquire specific information of video events by structured scene graph generation tool and descriptive image caption generation tool. The final response is achieved by incorporating the intrinsic world knowledge.}
  \Description{Enjoying the baseball game from the third-base
  seats. Ichiro Suzuki preparing to bat.}
  \label{fig:teaser}
\end{teaserfigure}


\maketitle

\pagestyle{plain} 

\renewcommand{\thefootnote}{\fnsymbol{footnote}}
\footnotetext[1]{Equal Contribution.}
\footnotetext[2]{Corresponding author.}
\renewcommand*{\thefootnote}{\arabic{footnote}}

\section{Introduction}

Video content comprehension, \emph{i.e.,} producing textual responses for videos based on the user instructions  remains a crucial and challenging topic, as it requires the mastery of skills for models including (1) visual content understanding from sparse videos and (2) multimodal reasoning with world knowledge.
With the variation in instruction types, this topic is divided into individual tasks, in which Video Question Answering (VideoQA)~\cite{zhong2022video} that focuses on answering user questions about videos, and Dense Video Captioning (DVC)~\cite{moctezuma2023video} that involves temporarily localizing and captioning all events in videos are two representative tasks\footnote{The Video Captioning task that usually accompanies by short clips is regarded as a specific case of DVC that only contains one event.}.

Existing studies solve these tasks by training video-language models on various video corpus, either by supervised learning based on domain-specific video-text annotations~\cite{wang2023learning,wang2021end,gao2023mist} or by unsupervised learning with massive plain videos~\cite{wang2023all,yang2023vid2seq,alayrac2022flamingo}.
However, due to the intrinsic redundancy of videos compared to images and texts, training efficiency remains a crucial problem, as the long videos may contain considerable amount of repetitive information with absence of knowledgeable texts.

\begin{figure}[t]
    \centering
    \includegraphics[width=0.48\textwidth]{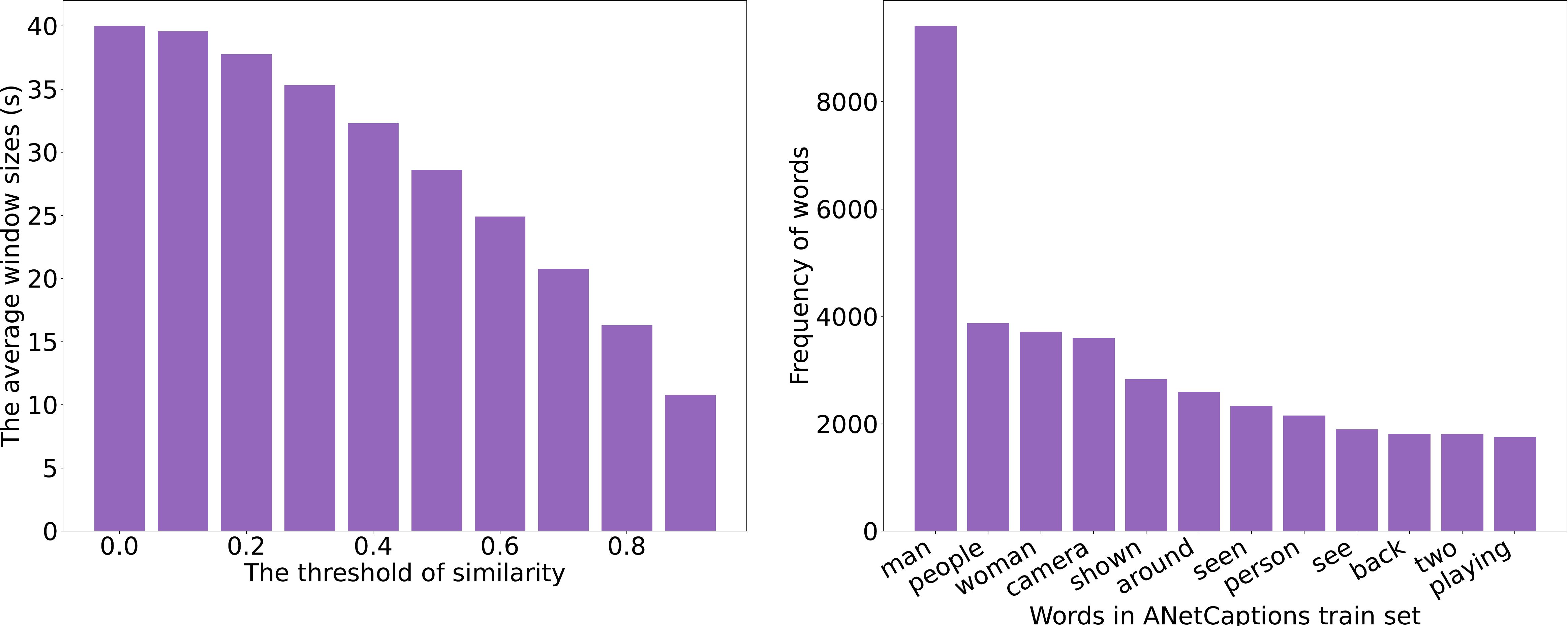}
    \caption{Average results on 100 randomly selected videos-captions from ActivityNet-Captions. Left: the counts of videos with average similarities of local frames. Right: the frequencies of words in captions.}
    \label{fig:example_stat}
\end{figure}

For example, popular video-language models (\emph{e.g.,} Flamingo~\cite{alayrac2022flamingo} and Vid2Seq~\cite{yang2023vid2seq}) would have to be trained on over 18 million videos using more than 1,000 ships of TPUv4 for 2 weeks to gain the knowledge reasoning capabilities.
An investigating on the randomly selected $100$ videos from ActivitNet-Captions dataset~\cite{krishna2017dense} is shown in Figure~\ref{fig:example_stat}.
We calculate the histogram-based similarity in various size of local windows of frames on time-axis, and statistics the word frequencies of corresponding captions.
We find that on average each frame has more than $16$ local frames with greater than 0.8 similarity score, and most of the captions merely contain rigid conceptual words instead of informative knowledge.

Meanwhile, Large Language Models~\cite{brown2020language,ouyang2022training,chowdhery2022palm,chung2022scaling,touvron2023llama} trained on web texts, such as Wikipedia, QA communities, Books and News have shown the remarkable knowledge reasoning and in-context learning abilities by only providing proper prompts with a few demonstrations. Built upon~\cite{ouyang2022training}, ChatGPT is one of the most representative LLMs that train a decoder-only Transformer~\cite{vaswani2017attention} using reinforcement learning from human feedback (RLHF)~\cite{christiano2017deep}.
Adopting LLMs as the reasoning agents offers the potential for the deployment of practical systems~\cite{zeng2022socratic,yao2022react,yang2023mm}.

However, a crucial challenge is the difficulty of language models to interact directly with videos, which prevents LLMs from using language instructions to concentrate on the most relevant visual content in videos. As shown in Figure~\ref{fig:teaser}, humans can answer the question $Q$ according to the event happened between $15.5$s and $151.8$s. The concentration is the key to facilities the accuracy and efficiency of reasoning from long videos.

In this paper, we introduce \textbf{VidCoM}, a fast tuning-free video comprehension framework based on large language model and lightweight multimodal tools.
Specifically, we first reveal that the key to successively achieve responses on long videos is the concentration on the most relevant video events, and the events information can be gathered and represented by two essential visual tools - the scene graph generation tool that generates structured information of objects with their detailed positions and relationships for visual scenes (\emph{e.g.,} \textit{Person$_2$[955,485,1060,672], behind, Car$_1$[1602,489,1713,604]}), and the image caption generation tool that provides descriptive information of existence and actions of images (\emph{e.g.,} \textit{Two skateboarders speed down a road passing by cars, tents and spectators.}) effectively.
Then, we propose to adopt the LLM equipped with dense world knowledge as the reasoning agent to perform multiple steps of detailed reasoning on the video to achieve the final response.
To address the challenge of enabling LLMs to perceive and attend to the most relevant video content, we further propose an Instruction-Oriented Video Events Recognition (InsOVER) algorithm based on efficient Hungarian matching.
By decomposing the language instruction and video event into OIE-triples and key-frames respectively, this algorithm can swiftly calculate the corresponding cross-modal similarity, enabling LLMs to specify the most relevant visual events using language instructions. The InsoVER algorithm bridges the gap between language models and video streams.

We conduct extensive experiments on two typical video comprehension tasks, namely Video Question Answering and Dense Video Captioning. The experimental results show that our method achieves state-of-the-art performance on both two benchmarks of STAR and ActivityNet-Captions, demonstrating the effectiveness.
Moreover, the proposed framework could be promoted flexibly by the further improvement of lightweight tools.

\section{Methodology}

The ultimate goal of video content comprehension is to build models that acquire textual responses to user instructions on video. To address the difficulties of training efficiency and knowledge reasoning, we build a framework that performs instruction-oriented multiple steps reasoning on videos based on the LLMs agents.
As the LLMs are not capable of perceiving visual signals, two image-level visual tools (\emph{i.e.,} scene graph generation and image caption generation) are employed to extract essential structured and descriptive information.

In this section, we first outline the problem of video content comprehension and our overall framework. And then, we describe the details of each module in our framework, including (1) the visual content acquisition using the lightweight multimodal tools, (2) the details of InsOVER algorithm for video event recognition, and (3) the process of employing LLMs to reason the final response on the gathered video events.

  \begin{figure*}
      \centering
      \includegraphics[width=\textwidth]{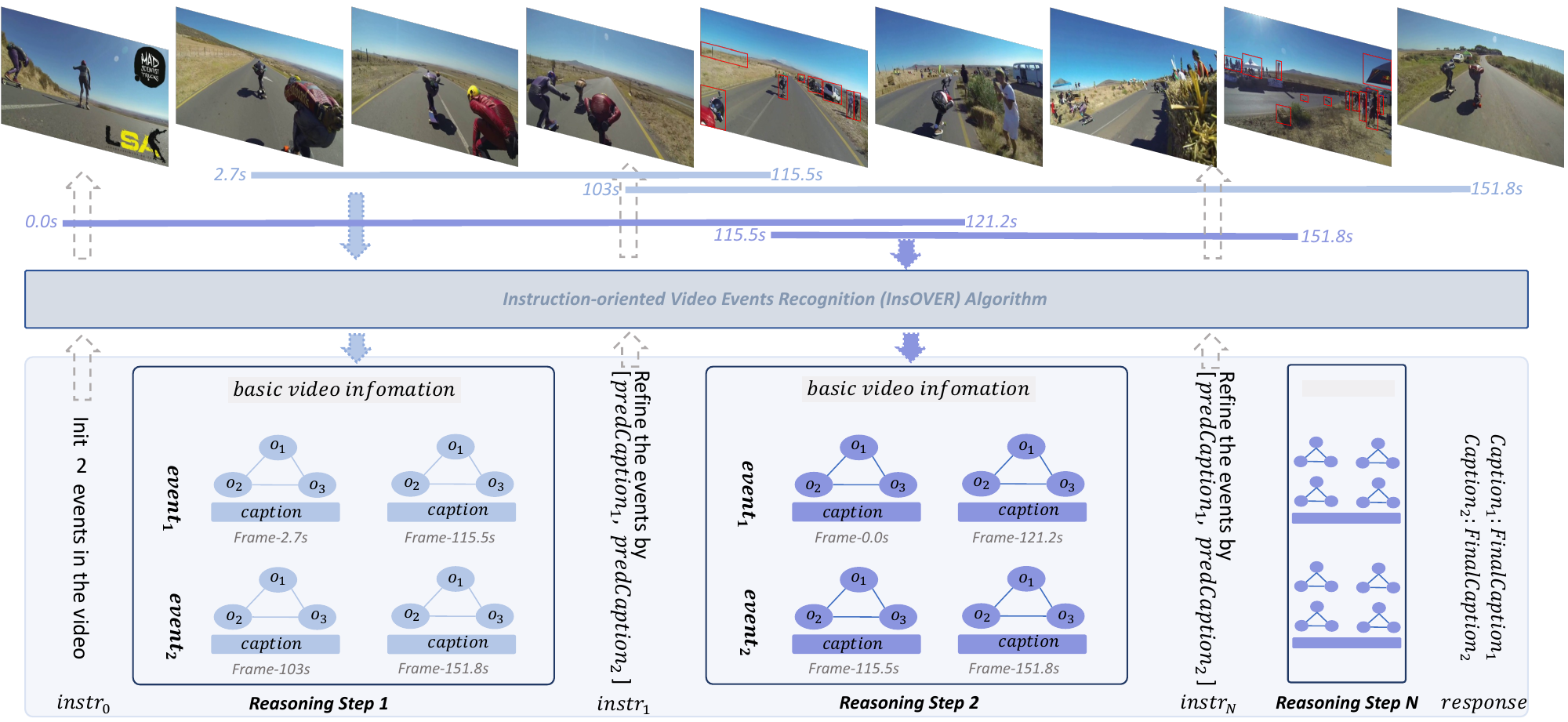}
      \caption{Illustration of the process of VidCoM with an DVC example. Given the user instruction requesting the events regions with captions, the InsOVER S-1 algorithm is adopted firstly to initialize $n$ events. A then $T$ reasoning steps of LLM agent on the video events are performed based on the InsOVER S-2 to achieve the final response.}
      \label{fig:model_main}
  \end{figure*}

\subsection{Overview}
\label{sec:overview}

We first give a formalized definition for the problem. Given a video $V=(E_1, E_2, ..., E_{n})$ consisting of $n$ visual events, the task of video content comprehension aims to build model $p_{\theta}$ that generates textual response $A$ to an user instruction $L$ on the video, based on the world knowledge $\mathcal{K}$,

  \begin{equation}
      p_{\theta}(A| V, L, \mathcal{K})
  \end{equation}

where each video event $E=(F_1, F_2, ...)$ refer to a temporally ordered sequence of frames, and the instruction $L$ and the response $A$ are two sequences of words. Distinguished by the types of instructions, this task has been divided into individual tasks, such as the task of Video Question Answering where the questions and answers serve as the instructions and responses, and the task of Dense Video Captioning where the fixed requirement and corresponding captions refer to the $I$ and $R$.
Existing studies mostly train video-language models $p_{\theta}$ on specific video-text annotation or massive plain video corpus, which suffer from the inefficiencies in training and deficiencies in knowledge acquisition.

The overall framework with an DVC example of \textbf{VidCoM} is illustrated in Figure~\ref{fig:model_main}.
Given a video that contains multiple video events, the process of VidCoM involves $T$ reasoning steps. First at the initial step, we adopt the proposed InsOVER algorithm to automatically initialize $n$ instruction-regardless video events (\emph{e.g.,} two events of \textit{[2.7, 103]} and \textit{[103, 115.5]}) from the video. Then, a sequence of $T$ interactive reasoning steps that interact LLM with the video based on InsOVER algorithm are performed to achieve the final predictions of repossess. For each interactive step, we (1) estimate the captions of the video events based on the crucial information (\emph{i.e.,} structured and descriptive information of scene graphs and captions) extracted from the key frames from each events as well as basic video information, and (2) refine the boundaries of each event by utilizing InsOVER algorithm based on the estimated captions. The interactive reasoning step is repeated until the changes in boundaries converge to a predefined constant $\delta$ or the maximum step $T$ is reached\footnote{Refer to the Sec.~\ref{sec:experiment} for the settings of $\delta$ and $T$ for VideoQA and DVC, respectively.}.
By performing these instruction-oriented reasoning steps incorporating the world-knowledge of the LLM agent, the process end up with the final predictions of the responses.
We introduce the details of the individual reasoning steps and the InsOVER algorithm in the following sections.

\subsection{Video Content Acquisition with Lightweight Multimodal Tools}
\label{sec:content_acquire_with_tools}

In order to get a specific response (\emph{e.g.,} answer to a visual question or caption for an video event) corresponding to an user instruction on a sparse video, it is sufficient by concentrating on the informative key frames that are most relevant to the instruction~\cite{gao2023mist}.
Scene graph generation (SGG) that extracts structured information of objects with their positions and spatial relationships and image caption generation (ICG) that provides descriptive information of existence and  activities in images have been proven effective on the video content comprehension tasks~\cite{pan2020spatio,cherian20222}.

In this study, we leverage two effective image-level models IETrans~\cite{zhang2022fine} and BLIP2~\cite{li2023blip} as the lightweight visual tools to extract scene graphs and captions from video frames, respectively.
Specifically, IETrans is a fine-grained scene graph generation model that can be applied in a plug-and-play fashion and expanded to large SGG with 1,807 predicate classes. We use the model with Motif~\cite{zellers2018neural} backbone trained on the VG-1800 dataset as the scene graph generation model $g^{SG}$.
BLIP2 is a image-language pre-training model built upon a framework consisting of two unimodal frozen models (\emph{i.e.,} an image encoder and a LLM) connected by a querying transformer with two-stages of training, where the first stage trains the vision-language representation with the frozen image encoder and the second stage trains the vision-to-language generation with the fronzen LLM.
We use the model implemented with the foundations of ViT-L/14 from CLIP~\cite{radford2021learning} and FlanT5$_{XL}$~\cite{chung2022scaling} as the image caption generation model $g^{IC}$.

Given the $i$-th frame $F_i$ in a video, we extract the scene graphs and captions based on the models:

  \begin{align}
  \begin{split}
      & \mathcal{G}_i = \{(s_i, r_{i}, o_i)\}_{i=1}^{N_g} = g^{SG}(F_i) \\
      & \mathcal{C}_i = (w_1, w_2, ..., w_{N_c}) = g^{IC}(F_i)
  \end{split}
  \end{align}

where $N_g$ and $N_c$ are the number of triples and and the number of words. The subject $s$ and object $o$ consists of an object bounding box $b\in\mathbb{R}^4$ and an object class $c$.
For the scene graphs, we find that there is a significant amount of inconsequential information (\emph{e.g.,} \textit{(head, on, shoulder)}) and some incorrect identifications. Therefore, we empirically filter out triples with confidence below a predefined threshold $\tau$\footnote{We set $\tau=0.4$ for IETrans in our work.}.
These two visual tools can further be alternated by adapting to a specific domain or the up to date versions.

The above visual information acquisition approach serves as the fast and effective atomic method which is adopted on particular frames in terms of language instructions.
In each reasoning step, VidCoM utilize the InsOVER algorithm to recognize video events with crucial frames that are specified by the language instructions of LLM agent, and then apply these atomic acquisition methods.

  \begin{figure}[pt]
      \centering
      \includegraphics[scale=0.36]{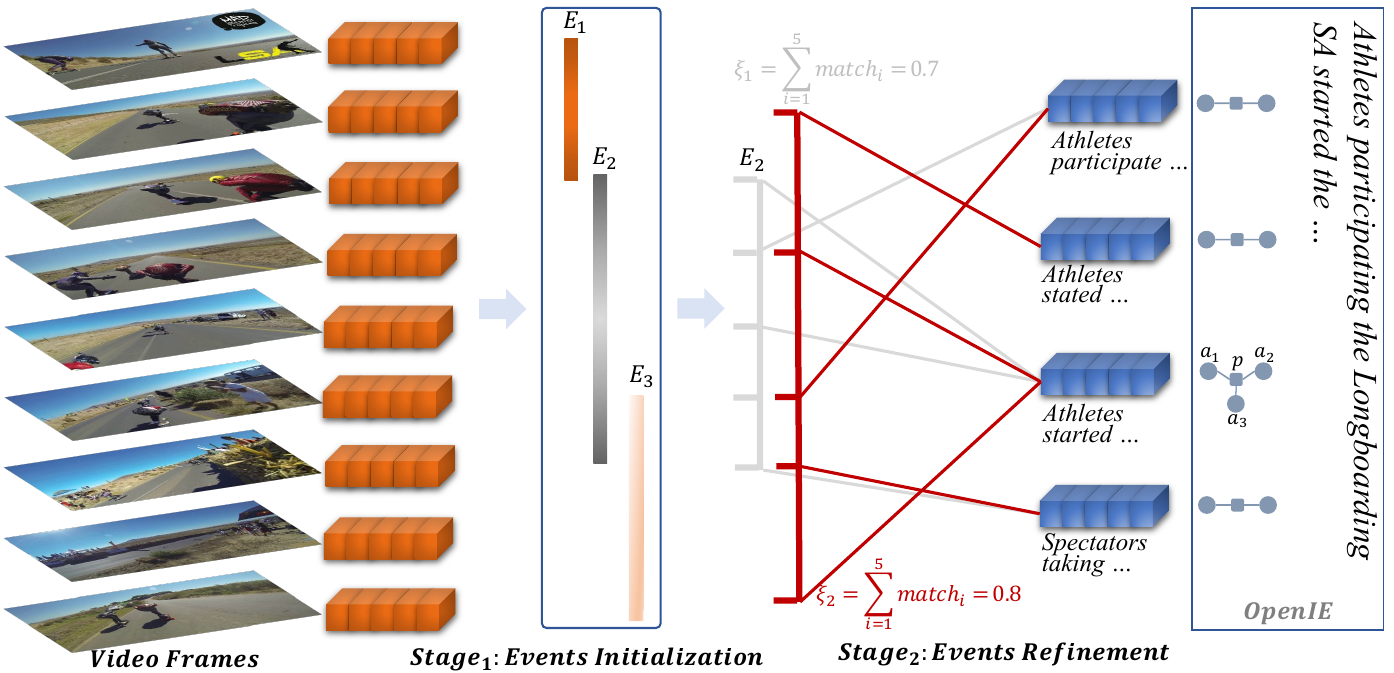}
      \caption{Illustration of the InsOVER algorithm, where the $stage_1$ initialize $3$ events automatically, and the $stage_2$ refine the events based on bipartite-graph matching between frames and assertions extracted from OpenIE model.}
      \label{fig:model_align}
  \end{figure}

\subsection{Instruction-Oriented Video Events Recognition (InsOVER)}
\label{sec:fast_algin_events}

The key to effective reasoning of LLMs on sparse videos is to recognize and concentrate on the most relevant video events~\cite{li2022invariant}.
However, there is an intrinsic difficulty that prevents LLMs from perceiving video content due to the modality limitation.
To enable LLMs to interact with videos using natural language instruction directly, we propose InsOVER, a fast and effective algorithm that recognizes the instruction-oriented videos events in a given video.

Given a long video with an uncertain content distribution,  the proposed algorithm consists of two sequential stages: (1) at the first stage S-1, a moving average-based linear clustering approach along the time axis is employed to yield $n$ event boundaries coarsely, and (2) at the second stage S-2: a Hungarian algorithm-based bipartite graph matching between decompositions of the language instruction and video event is performed to refine the events boundaries meticulously.
Both stages guarantee the low time complexities on videos.

\subsubsection{\textbf{S-1: Moving Average-based Events Initialization}}
This stage aims to automatically initialize $n$ events with their boundaries in a video, where $n$ is pre-specified according to specific tasks.
Specifically, given a video $V$, we first perform an uniform sampling on the frames of the video with a constant frequency (\emph{e.g.,} 2 frames/second), resulting frames $(F_1, F_2, ...)$. Then, we encode these frames into a sequence of hidden representations $(\mathbf{h}^F_1, \mathbf{h}^F_2, ...)$ by using a pre-trained visual encoder ViT-L/14 from CLIP and acquiring the first embedding vector of the special token of each patched sequence of visual tokens. These representations reflect the features of major visual contents of the video frames. Next, we uniformly initialize $n$ indices of frames $\{i | i = 1, 2, ..., n\}$ excluding the start and end frames as the initial central points of the events $\{R^i_{b:e}\}$\footnote{The $b$ and $e$ refer to the beginning and end indices of frames.}, and perform multiple epochs of expansion processes for each event.

For $i^{th}$ event, at each epoch, we first calculate the average representation of frames in current region $R^i_{b:e}$, and then iteratively expand the region by comparing the similarity between the representation of $j^{th}$ left/right frame $F_{b-j}/F_{e+j}$ and the average representation:

\begin{align}
    sim(F_j, [F_b, ..., F_e]) = cos(\mathbf{h}^F_j, \frac{1}{b-e+1}\sum_{k=b}^e \mathbf{h}^F_k)
\end{align}

where the average representation is calculated once at the beginning of each epoch to ensure the stability and efficiency. The current region will be expanded to incorporate the new frame if the similarity score is greater than or equals to a constant threshold $\delta$, and the expansion process is continued until the condition is not satisfied\footnote{We empirically set the value of $\delta$ to 0.95 when adopting ViT of CLIP in this work.}.
We parallel the epochs for all events until the regions do not change, and finally obtain the $n$ video events $(E^1_1, E^1_2, ..., E^1_n)$ after this stage.
In comparison to the conventional histogram-based methods~\citep{tang2019fast}, we empirically find that the sub-algorithm of this stage can be used to get video events for an arbitrary video precisely while preserving the efficiency.

\subsubsection{\textbf{S-2: Instruction-specified Events Refinement}} This stage aims to refine the boundaries of given events $(E_1, E_2, ..., E_n)$  meticulously according to specified language instructions $(L_1, L_2, ..., L_n)$, for enabling the linguistic manipulation on videos.
For each language instruction $L_j$, this problem can be formulated as finding a new event $E'_i\rightarrow R_{b:e}$ starting from $E_i$ where the new event is best aligned to the instruction $L_j$.
Roughly align a sentence of language instruction and a clip of video event is difficult and ambiguous.
We first present that each video event $E_i$ and language instruction $L_j$ can be further decomposed into multiple textual sub-events $L_j=(l^j_1, l^j_2, ..., l^j_{m_l})$ and visual sub-events $E_i=(e^i_1, e^i_2, ..., e^i_{m_v})$, respectively, where each sub-event refers to an specific predicate or action (\emph{e.g.,} the language instruction \textit{A participating in B started C} can be decompose into \textit{A participating B} and \textit{A started C}).
Then, the problem can be transformed to a bipartite-graph matching between $m_l$ textual sub-events of $\{L_i\}$ and $m_v$ visual sub-events of $\{E_i\}$, and the optimal matching score can be solved by the Hungarian algorithm efficiently.
We utilize the OpenIE model and key frames to obtain the textual and visual sub-events, and the cross-modal similarity to calculate the pairwise matching scores.

Specifically, for each pair of an video event $E^1_i$ that is initialized from the first stage and a language instruction $L_i$, we first adopt the RobustOIE~\citep{qi2022syntactically} model to obtain $m_l$ tuples $\{(a_1,p,a_2,...)_u\}_{u=1}^{m_l}$ consisting of arguments and predicates, and concatenate each tuple into a sentence of assertion as the corresponding textual sub-event $l^i_u$. Based on these textual sub-events, we iteratively change the current event region and calculate the bipartite-graph matching score to the visual sub-events obtained from the region at current iteration.
At $t^{th}$ iteration, we uniformly sample $m_v$ key frames along the time-axis including the boundaries (\emph{i.e.,} the $b$ and $e$) in $R^i_{b:e}$ as the visual sub-events $(e^i_1, ..., e^i_{m_v})$, and calculate the matching score:

\begin{align}
    &\xi^i_t = \sum_{(u,q)\in\psi} sim(l^i_u, e^i_q) \\
    &sim(l^j_u, e^i_q) = cos(\mathbf{h}^{li}_u, \mathbf{h}^{ei}_q)
\label{eq:graph_matching}
\end{align}

where the set of pairwise matchings $\psi$ are solved by the Hungarian algorithm. We utilize the textual encoder BERT and visual encoder ViT both from the multimodal model BLIP2 to encode $l^i_u$ and $e^i_q$ to obtain the corresponding hidden representations $\mathbf{h}^{li}_u$ and $\mathbf{h}^{ei}_q$, respectively.
We repeat the above iteration along the left-side firstly and the right-side secondly with the stride size $5$ for each boundary of event $E^i\rightarrow R^i_{b:e}$ resulting four trajectories, where each trajectory of searching is ended when the matching score $\xi_t$ do not increase.
By paralleling the searching trajectories on all events, the instruction-oriented video events $(E^2_1, E^2_2, ..., E^2_n)$ are achieved.
Based on the fine-grained matching and the Hungarian algorithm, this sub-algorithm guarantees both the accuracy and efficiency for language-video alignment.

\subsubsection{\bfseries Time Complexity} InsOVER-S1: the general time-complexity of moving-average is $\mathcal{O}(N)$. We empirically found that the algorithm converges quickly when we set a appropriate threshold based on a prior observation with BLIP2 and cosine similarity calculation. InsOVER-S2: as the time complexity of the Hungarian algorithm is $\mathcal{O}(nm)$, and the time complexity of our stride sampling is less than $n$, the final estimated upper bound of the time complexity is $\mathcal{O}(Nnm)$, where $N$ is the total number of frames, and $n$ and $m$ are the number of triplets and the number of frames (e.g., $n=3, m=5$).

\subsection{Textual Response Reasoning with LLMs}
\label{sec:responese_reason_with_llms}

A lot of systematic commonsense knowledge and empirical knowledge are not explicitly exhibited in visual scenes, and the indeed understanding and precise response generation require the mastery of these world knowledge~\cite{jin2023knowledge,gu2023text}.
Based on the InsOVER algorithm with visual tools, we present to achieve the language response to specific user instructions by a LLM reasoning agent that is equipped with world knowledge from large-scale pre-training.

Given a video $V$ with an user instruction $L^0$ that requests a corresponding response to the video, we first adopt the InsOVER S-1 to initialize $n$ video events with their region boundaries $(E^1_1, E^1_2, ..., E^1_n)$.
And then for each individual event $E^1_i\rightarrow R^1_{b:e}$, we perform a sequence of $T$ reasoning steps that adopt LLM as the reasoning agent to interact with the video based on the InsOVER S-2 to achieve the final response. At $t^{th}$ step, the LLM agent output the instruction $L^t_i$ based on current event information and basic video information:

\begin{align}
    L^t_i = f^{LLM}(info(E^t_i), info(V)) \\
    info(E^t_i) = \begin{cases}
        \{\mathcal{G}^t_j | F_j\in E^t_i\},  \\
        \{\mathcal{C}^t_j | F_j\in E^t_i\}, \\
    \end{cases}
\end{align}

where $\mathcal{G}^t_j$ and $\mathcal{C}^t_j$ refer to the scene graph and caption of $j^th$ frame sampled from the event $E^t_i$ respectively, and $info(V)$ is the basic information of the video (\emph{i.e.,} the duration and frame resolution) told to the agent.
Next, the refined event region $E^{t+1}_i$ is obtained by adopting the InsOVER S-2 on the the instruction $L^t_i$ and current event $E^t_i$.
The reasoning step is repeated until the changes in boundaries converge to a predefined constant $\delta$ or the maximum step $T$ is reached.
We parallel these reasoning steps for all initial events to acquire the converged video events $\{E^T_i\}$.
Finally, For the video captioning tasks, the response is obtained straightforwardly with the $T^{th}$ predicted instruction $L^T$. For the question answering tasks, we adopt the LLM to predict the corresponding answer based on the information of acquired events $\{info(E^T_i)\}$.
Detailed prompts we designed in this study are available at Appendix Sec. 3.

\section{Experiments}
\label{sec:experiment}

To demonstrate the effectiveness of the proposed framework VidCoM, we conduct extensive experiments on two typical video content comprehension tasks, namely Video Question Answering (VideoQA) and Dense Video Captioning (DVC).
We experiment the implementation of LLM agent in this study with two representative models: the close-source model ChatGPT and open-source model LLaMA2~\citep{touvron2023llama}, where the former is one of the currently best-performing proprietary models and the latter is a readily expandable open-source foundational model.

\subsection{Experiments on Video Question Answering}

Given a question $Q$ asking about the specific content of a video, the task of VideoQA aims to generate language answer $A$ in for the question. The $Q$ and $A$ are corresponding to the instruction $L^0$ and final response $A$ for adapting into our framework naturally.

\begin{table*}[ht]
\renewcommand{\arraystretch}{1.1}
\centering
    \begin{tabular}{p{2cm}|p{2.8cm}|c|ccccc}
    \toprule
    \multirow{2}{*}{\textbf{Supervision}} & \multirow{2}{*}{\textbf{Model}} & \multirow{2}{*}{\textbf{Training Modality}} & \multicolumn{5}{c}{\textbf{Question Type}}                                                                                                                          \\ \cline{4-8}
                                 &                        &                   & Int\_Acc (↑) & Seq\_Acc (↑) & Pre\_Acc (↑) & Fea\_Acc (↑) & Mean  \\ \hline
    \multirow{4}{*}{\textit{Supervised}}  & SHG-VQA\citep{urooj2023learning} & Video-Text        &      47.98   &    42.03     &     35.34     &   32.52    &   39.47$^\dagger$    \\
                                 & MIST\citep{gao2023mist}  & Video-Text        & 55.59        & 54.23        & 54.24        & 44.48        & 51.23 \\
                                 & InternVideo\citep{wang2022internvideo}  & Video-Text   &  62.7  &  65.6  &  54.9  &  51.9  & 58.7      \\
                                 & SEVILA\citep{yu2023self}  & Video-Text   &  63.7  &  70.4  &  63.1  &  62.4  & 64.9      \\ \hline
    \multirow{4}{*}{\textit{Few-shot}}     & Flamingo-80B\citep{alayrac2022flamingo}& Video-Image-Text                   & 42.15       & 44.56       & 40.64   & 41.57   & 42.23$^\dagger$ \\
                                 & Flamingo-9B\citep{alayrac2022flamingo}  & Video-Image-Text                   & -       & -      & - & -    & 43.4 \\
                                 & SEVILA \citep{yu2023self}      & Video-Image-Text                 & 48.3    & 45.0 & \textbf{44.4} & 40.8    & 44.6 \\
                                 & \textbf{VidCoM}$_{llama2}$               & Image-Text                         & 28 & 48 & 28 & 20 & 31 \\
                                 & \textbf{VidCoM}$_{chatgpt}$               & Image-Text                         &    \textbf{52}   &    \textbf{52}   &    32   &     \textbf{64}  & \textbf{50}    \\ \hline
    \end{tabular}
\caption{The VideoQA performance on the STAR validation set, $\dagger$ indicates result on test set. The standard versions of VidCoM$_{llama2}$ and VidCoM$_{chatgpt}$ are two implementations based on the LLaMA2-chat-hf with 2-shot and 4-frames and the ChatGPT with 6-shot and 4-frames, respectively.}
\label{tab:vqa-main}
\end{table*}

\begin{table}[ht]
\renewcommand{\arraystretch}{1.1}
\centering
\begin{tabular}{l|ccccc}
    \toprule
    \multirow{2}{*}{\begin{minipage}{1.4cm}\textbf{Refine}\\\textbf{Iterations}\end{minipage}}   & \multicolumn{4}{c}{\textbf{Question Type}}     \\ \cline{2-6}
                     & Int\_Acc & Seq\_ Acc & Pre\_Acc & Fea\_Acc & Mean\\ \hline
    ${T=0}$ &    48   &    32   &    24   &     36  & 35\\
    ${T=1}$ &    48   &    48   &    36   &     48  & 45\\
    ${T=2}$ &    44   &    40   &    44   &     40  & 42 \\
    ${T=3}$ &    36   &    56   &    32   &     32  & 39 \\
    \hline
    \end{tabular}
\caption{Ablation studies with various number of refinement iterations on STAR.}
\label{tab:vqa-refine}
\end{table}

\subsubsection{\textbf{Experiment Settings}}

To examine the performance of our method on video question answering tasks, we adopt the newly proposed challenging and representative dataset, STAR~\citep{wu2021star} as evaluation benchmark. STAR composes of about 9,000 videos sampled from Charades dataset \citep{sigurdsson2017actions} and most of the videos depict indoor human activities.
For each video, a question with a correct answer from four candidates are annotated by human works. The questions are categorized into four types, namely Interaction, Sequence, Prediction and Feasibility, requiring various types of reasoning abilities.
In light of substantial time and financial costs, we sampled 25 questions of each categories from the original validation set as our test set and report the accuracy score on each category as well as the average.

We experiment with different hyper-parameters settings for VidCoM, which varying the employment of LLM agent, the number of frames sampled for each video event (\emph{i.e.,} $N_F$ frames), the the number of demonstrations prepended to the LLM agent (\emph{i.e.,} $N_D$ shots), and number of refinement iterations (\emph{i.e.,} $T$ iterations) for an exhaustive evaluation. We empirically refer to the ChatGPT-based model with the optimal settings of $N_f=4, N_D=6, T=1$, and the LLaMA-based model with the optimal settings of $N_f=4, N_D=2, T=1$ as the standard implementations of VidCoM$_{chatgpt}$ and  VidCoM$_{llama2}$, respectively.
We use the ChatGPT model based on the official API of OpenAI~\footnote{The experiment period of calling \href{https://platform.openai.com/docs/models/gpt-3-5}{the API} is from May 12, 2023, to December 9, 2023.}, and the LLaMA2-13b-chat-hf as our deployment.
To regulate the output format and enhance the ability, we randomly pick two videos from train set and use our deployed tools to generate the corresponding text-based information, which, together with the question and correct answer, serves as the demonstration for LLM.

\subsubsection{\textbf{Experiment Results}}

The experiment results in comparison with our standard models against existing SOTAs are shown in Table~\ref{tab:vqa-main}, where the supervised methods were trained on the training set of STAR in advance, and the fow-shot methods employ a few annotated examples as demonstrations to prepend on the inputs.
Overall, we can see that our model achieves the superior performance across the board of few-shot setting. Our standard model VidCoM$_{chatgpt}$ outperforms the previous best method by $4.4$ percentage points resulting a state-of-the-art performance.
In comparison to the existing supervised methods, our model also exhibits a compatible performance by only adopting $6$ demonstrations compared to the full training set of thousands of videos.
These results suggest that VidCoM$_{chatgpt}$ can address the general video task of VideoQA effectively in a tuning-free manner, by adopting LLMs to reasoning on videos based on the proposed InsOVER algorithm.
Particularly, VidCoM$_chatgpt$ obtains the maximum accuracy score of $64$ on the split of question type of Feasibility compared to the $41.57$ of Flamingo-80B, where the latter performs massive training on 18 million videos and 80 billion parameters. This result suggests that our model can understand and incorporate the commonsense effectively by benefiting from the dense world-knowledge of LLMs, and make reliable decisions.

\begin{figure}[t]
    \centering
    \includegraphics[width=0.48\textwidth]{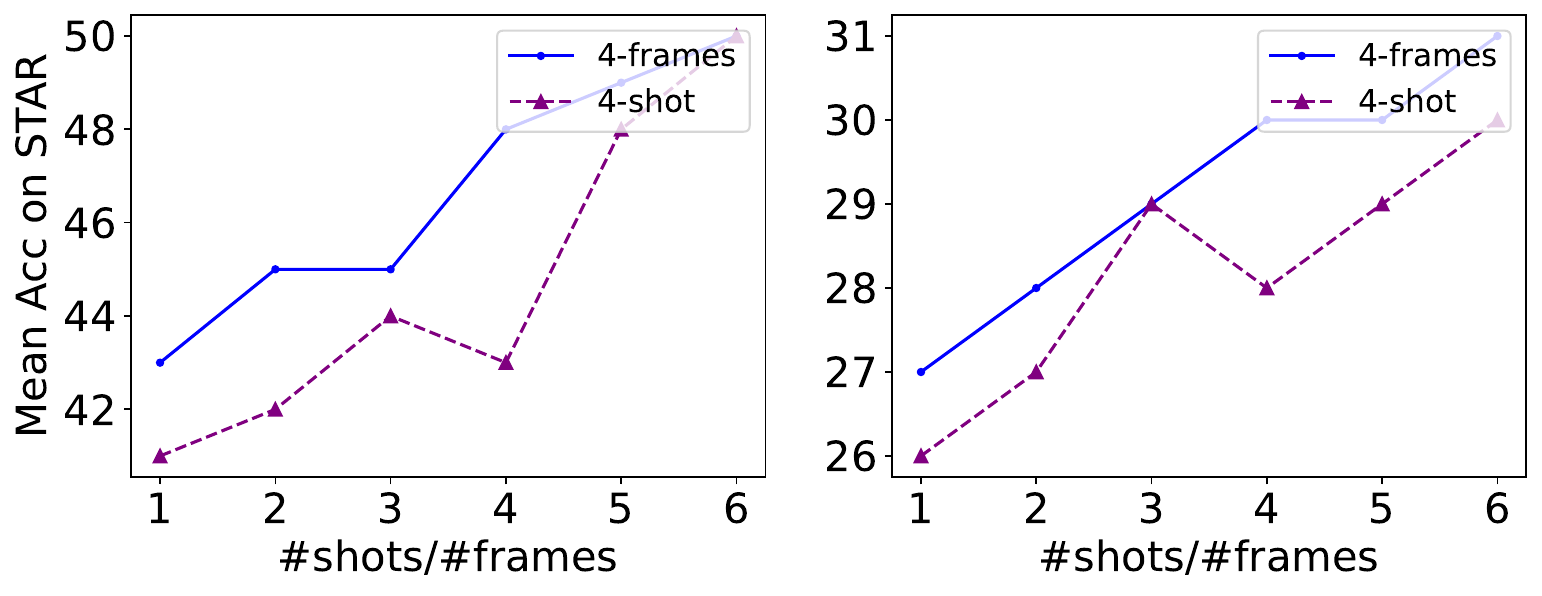}
    \caption{Ablation studies with various numbers of demonstrations and frames on STAR.}
    \label{fig:ablation_on_star}
\end{figure}

\begin{figure*}[pt]
    \centering
    \includegraphics[width=\textwidth]{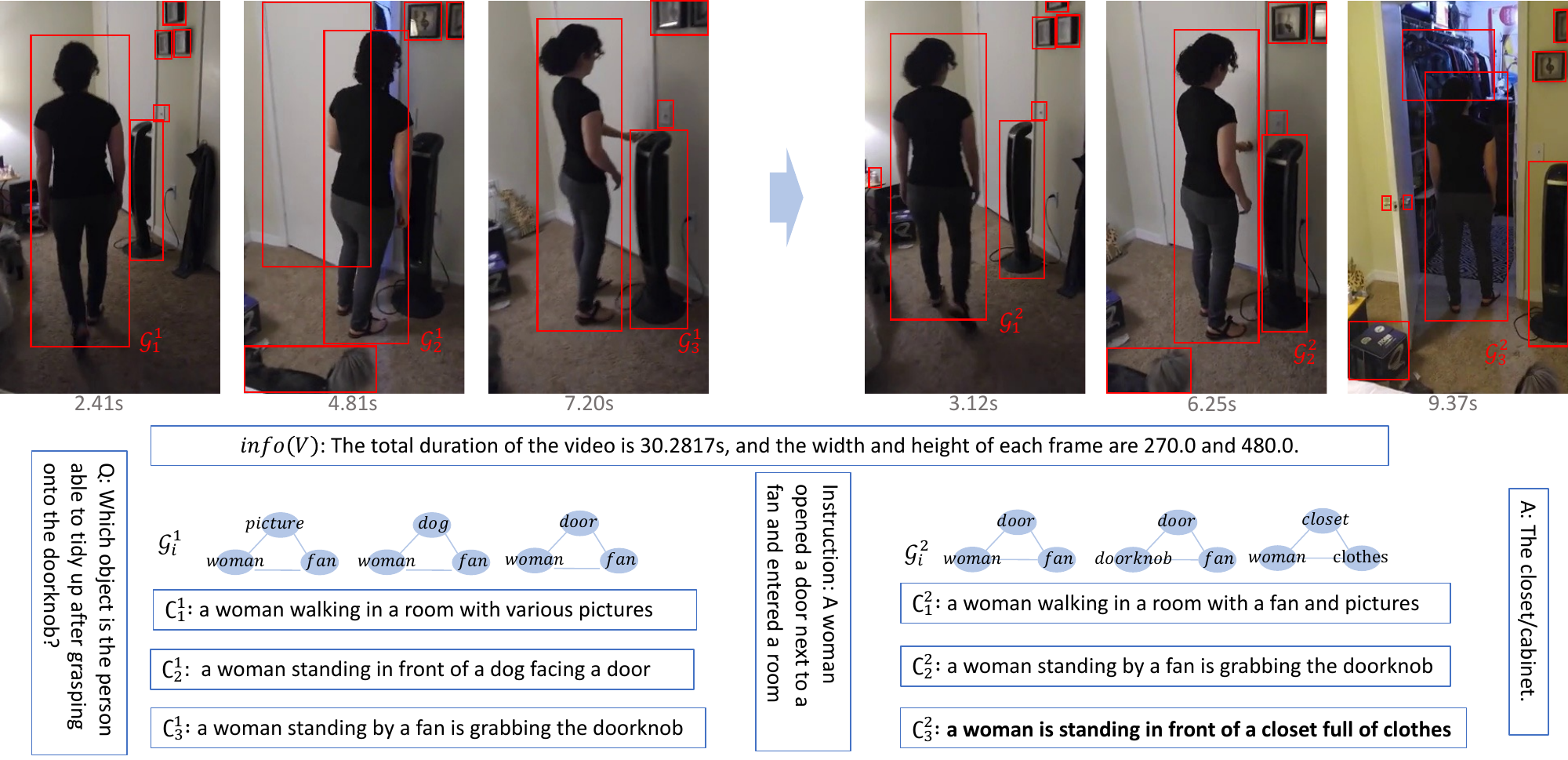}
    \caption{A case study of VidCoM on STAR.}
    \label{fig:case_star}
\end{figure*}

\subsubsection{\textbf{Ablation Study}}

In order to investigate the impact on different settings of major hyper-parameters, we further conduct detailed ablation studies in varying the number of refinement iterations, the number of sampled frames in each event, and the number of demonstrations provided into LLM agents.
The results of VidCoM$_{chatgpt}$ with various refinement iterations $T$ on STAR benchmark are shown in Table~\ref{tab:vqa-refine}. From the results we can see that the instruction-oriented refinement S-2 significantly improve the overall reasoning performance by a margin of $10\%$, compared to only using S-1 on InsOVER.
It demonstrates that the LLM agent could interact with the video effectively through the S-2 of InsOVER algorithm, and directly promote the success of task.
Moreover, we can also notice that compared to the optimal result, the performance decreases when increase the refinement iterations to exceed 2. This phenomenon implies that based on the initialization of S-1, the LLM agent can refine the corresponding event boundaries efficiently in the early prediction of S-2, while excessive refinement may cause the event content to drift relative to the initial user instruction.
Figure~\ref{fig:ablation_on_star} shows the performance of VidCoM$_{chatgpt}$ by varying the number of frames selected in each event and the number of demonstrations provided to the LLM agent.
We can see that the overall performance of model variants with the fixing of 4 frames is better than the variants with fixed 4 shots for both implementations. This suggests that the number of frames in obtained each event is substantially importance in VidCoM.
In additional, we observe that VidCoM$_{chatgpt}$ and VidCoM$_{chatgpt}$ achieve the optimal performance with 4 frames while fixing the number of demonstrations to $4$.

\subsubsection{\textbf{Case Study}}

Figure~\ref{fig:case_star} showcase the reasoning process of VidCoM on an example of STAR with detailed information of video events extracted from visual tools and linguistic instructions generated from LLM agent.
Given a question of \textit{Which object is the person able to tidy up after grasping}, the framework first adopt InsOVER S-1 to initialize one event with the start and end boundaries of $2.41s$ and $7.20s$. Then, two visual tools (\emph{i.e.,} scene graph generation and image caption generation) are used to extract the event information of corresponding scene graphs $(\mathcal{G}^1_1, \mathcal{G}^1_2, \mathcal{G}^1_3)$ and captions $(\mathcal{C}^1_1, \mathcal{C}^1_2, \mathcal{C}^1_3)$ from the uniformly sampled 3 frames, respectively.
By incorporating the event information and basic video information $info(V)$, the LLM agent predict the instruction of \textit{A person able to tidy up after grasping onto the doorknob} that will be used to refine the current event boundaries.
By performing the InsOVER S-2 based on the LLM instruction, the new event boundaries $3.12s$ and $9.37s$ are obtained accordingly, and the same video event acquisition process is performed to obtain the updated scene graphs $(\mathcal{G}^1_1, \mathcal{G}^1_2, \mathcal{G}^1_3)$ and captions $(\mathcal{C}^2_1, \mathcal{C}^2_2, \mathcal{C}^2_3)$.
Finally, the LLM agent is able to predict the answer of \textit{The closet/cabinet} conveniently based on the question-specified video event information.

\subsection{Experiments on Dense Video Captioning}

In order to investigate the general applicability of VidCoM, we further conduct experiments on anther typical video content comprehension task, Dense Video Captioning (DVC).
DVC is a substantial challenging task that aims to localize the start-end boundaries of multiple video events and generate the corresponding captions in a given video, where the subjectivity of video event boundaries and the freedom of description both make this task extremely difficult for current vision-language models.

\subsubsection{\textbf{Experiment Settings}}

We use the standard DVC benchmark, ActivityNet-Captions\citep{heilbron2015activitynet} for our evaluation. ActivityNet-Captions is a widely-used video-text benchmark that contains 20K untrimmed videos with dense event annotations, where each video is annotated with $3.7$ temporally-localized captions.
For sake of efficiency, we randomly sample 50 videos from the original validation set as our evaluation set.
For the evaluation metric, we adopt the de facto benchmark of Story Oriented Dense Video Captioning Evaluation Framework (SODA) to validate the performance. In comparison with traditional matrices (\emph{e.g.,} BLEU ro METEOR), SODA finds temporally optimal matching between generated and reference captions to capture the story of a video, which fairly reflects the performance of DVC models.
Similar as the experiment on VideoQA, we also implement the agent in VidCoM with two LLMs of the same versions, ChatGPT and LLaMA2 for a sufficient validation.
We perform a thorough study to find the optimal hyper-parameters, including the number of frames in each video event $N_F$, the number of demonstrations provided to the LLMs $N_D$, and the number of refinement iterations $T$.

\begin{small}
    \begin{table}[ht]
    \centering
    \begin{tabular}{p{1.15cm}|l|c|c|c}
    \toprule
    \textbf{Category} & \textbf{Model} & \begin{minipage}{1cm}\textbf{Train\\Modality}\end{minipage} & \begin{minipage}{1.1cm}\textbf{Pred\\proposals}\end{minipage}    & \begin{minipage}{1cm}\textbf{True\\proposals}\end{minipage}     \\ \hline
    \multirow{2}{*}{\textit{Supervised}}  & PDVC\citep{wang2021end}    & Video-Text  &   5.4  & - \\
     & Vid2Seq\citep{yang2023vid2seq}   & Video-Text   & 5.8  & - \\ \hline
    \multirow{3}{*}{\textit{Few-shot}} & Vid2Seq~\citep{yang2023vid2seq}  &   Video-Text   &  2.2  & - \\
    & \textbf{VidCoM}$_{llama2}$ &  Image-Text & 3.3  &  4.9   \\
    & \textbf{VidCoM}$_{chatgpt}$ &  Image-Text &  \textbf{3.7}  &  6.6   \\ \hline
    \end{tabular}
\caption{DVC results on ANet val-set. All implementations adopt $1$ refinement step, and use the settings of $4/4$ frames/shots and $4/2$ frames/shots, respectively.}
\label{tab:anet-main}
\end{table}
\end{small}

\subsubsection{\textbf{Experiment Results}}

The experimental results of our optimal implementations in comparison to previous SOTAs of supervised and few-shot categories are shown in Table~\ref{tab:anet-main}.
We empirically find the optimal settings of hyper-parameters as $N_F=4, N_D=4,T=1$ and $N_F=4, N_D=2,T=1$ for VidCoM$_{chatgpt}$ and VidCoM$_{llama2}$, respectively.
Overall, our model surpass the previous best-performing model Vid2Seq by $1.5$ SODA score achieving the state-of-the-art perfomrance in few-shot setting, where the latter is trained on a massive of 18 million videos.
Compared to the fully-supervised methods trained on the training set of 10K videos of ActivityNet-Captions, our model also obtains the compatible performance with existing SOTAs.
In order to investigate the impact of the event boundaries on the captioning performance, we additionally conduct experiments by proving the golden event boundaries (\emph{i.e.,} True proposals) to the LLM agents to perform the predictions of captions directly.
As shown in the result, based on the correct boundaries of events, our model further improve the performance to a SODA socre of $6.6$, which achieves the state-of-the-art performance across the board event outperforming the previous supervised SOTA model.

\section{Related Work}
\subsection{Video Content Comprehension}

There is a long line of works to fulfill the topic of texts generation based on given videos, where the Video Question Answering (VideoQA)~\cite{zhong2022video} and Dence Video Captioning (DVC)~\citep{moctezuma2023video} are two representative tasks that have been drawn growing attention.
Traditional VideoQA studies train the dedicated models based on LSTM and graph neural networks to capture the cross-modal interaction~\citep{zhao2018open,li2019beyond,park2021bridge} or motion-appearance information~\citep{gao2018motion,le2020hierarchical}. Further, the Transformer-based architectures have shown to be effective in modeling the multimodal fusions~\citep{xiao2022video}. With the success the large-scale pre-training on multimodal domain~\citep{li2022align,zhong2023stoa, wang2023all,ye2023hitea,ma2023temporal}, the models with substantial amount of parameters trained on massive videos are achieve remarkable performance on this task~\citep{bain2021frozen,fu2021violet,buch2022revisiting,alayrac2022flamingo}. Recently, adopting models to acquiesce answers based on the crucial segments of frames are demonstrated the accuracy and time efficiency\citep{lei2021less,gao2023mist,yu2023self}.

Dence Video Captioning is a challenging and practical task that has been studied for many years~\citep{xiankai2023survey}.
Starting from the first DVC model~\citep{krishna2017dense} that adopts LSTM as the text generation model with a multi-scale proposal module for video events localization, early works on this task mainly focus on the modeling of multimodal-contexts~\citep{wang2018bidirectional,yang2018hierarchical} or event-level relationships~\citep{iashin2020multi,iashin2020better} with the similar generation architecture.
To address the limitation of lacking interaction between generation module and localization module, some studies propose to using multi-objectives optimization~\citep{li2018jointly} or the masking mechanism to link the gradient flow from captioning loss to proposals’ boundaries~\cite{zhou2018end}.
Furthermore, the PDVC model~\cite{wang2021end} treat the task as a set prediction, and jointly perform event localization and captioning for each event in parallel.
Recently, the video-language pre-trained models have been explored and applied to the temporal localization tasks~\cite{lei2021detecting,wang2022contrastive,xu2022contrastive,yang2021taco,yang2023vid2seq,zellers2022merlot}.

\subsection{Large Language Models for Multimodal Generation}

To enable large language models better understanding visual information and generating natural language, there are two categories of works that adopt different solutions currently. The first category aims to project visual information into the space of large language models based on the pret-rained visual encoders~\citep{hu2022promptcap,wang2022language,liu2023visual,zhu2023minigpt,dai2023instructblip}. For example, the MiniGPT-4~\citep{zhu2023minigpt} aligns a frozen visual encoder with a frozen LLM, Vicuna, using just one projection layer and achieves considerable abilities including detailed image description generation and website creation. The LLaVA~\citep{liu2023visual} use language-only GPT-4 to  generate multimodal language-image instruction-following data and trains a large multimodal model that connects a vision encoder and LLM for general- purpose visual and language understanding. The InstructBLIP~\citep{dai2023instructblip} retains Q-Former which is similar as BLIP2~\citep{li2023blip}, and replace language model to a larger one, and then tuning on meticulously collected instruction data.

Another category of studies concentrate on adapting the large language models to the vison-based language generation problems in a tuning-free manner based on the flexible visual foundation models\citep{shen2023hugginggpt,yang2023mm, zeng2022socratic,wu2023visual}. For example, VisualGPT~\citep{wu2023visual} incorporates multiple visual foundation models to enable the user to interact with ChatGPT by sending and receiving information about response and images. The MM-REACT extend the REACT~\citep{yao2022react} model to the multimodal tasks with a pool of vision experts to achieve multimodal reasoning and action. Equipped with the extra tools, LLMs has proved to be eligible to solve this tasks. Another representative framework is the the Socratic Models~\citep{zeng2022socratic}, which adopts a modular framework in which multiple pretrained models may be composed zero-shot i.e., via multimodal-informed prompting, to exchange information with each other and capture new multimodal capabilities, without requiring finetuning.

\section{Conclusion}

In this paper, we present a tuning-free framework, \textbf{VidCoM}, a fast adaptive framework that leverage Large Langauge Models to comprehend and reason about videos with lightweight multimodal tools.
Specifically, we first reveal the key to response specific user instructions is the concentration on the most relevant video events, and then utilize the structured scene graph generation and descriptive image caption generation tools to gather and represent the corresponding video events.
A LLM equipped with world knowledge is then adopted as the reasoning agent to achieve the final response by performing multiple reasoning steps on the video events.
To address the difficulty of LLMs identifying video events, we further propose an Instruction-oriented Video Events Recognition (InsOVER) algorithm based on the efficient Hungarian matching. This algorithm localize the corresponding video events by calculating the similarity between decompositions of the linguistic instruction (\emph{i.e.,} into OIE-triples) and video event (\emph{i.e.,} key-frames), thus enabling LLMs to efficiently interact with long videos.
Experiments show that the proposed VidCoM outperforms heavily pre-trained models on typical video content comprehension tasks to achieve the state-of-the-art performance.


\bibliographystyle{ACM-Reference-Format}
\bibliography{www2024}

\appendix

\section{Details of InsOVER}
\subsection{Pseudo Code of InsOVER}

\begin{small}
    \begin{algorithm}[hpt]
\renewcommand{\algorithmicrequire}{\textbf{Input:}}
\renewcommand{\algorithmicensure}{\textbf{Output:}}
\caption{\textbf{InsOVER}}\label{alg:insover-s1}
\begin{algorithmic}[1]
\Require A video $V$, a threshold $\delta_1$
\Ensure $n$ video events $(E_1,E_2,..,E_n)$
\State Uniformly sample frames $(F_1,F_2,...)$ with FPS=1
\State Get hidden representations $(\mathbf{h}_1, \mathbf{h}_2^F, ..)=ViT(F_1, F_2, ..)$
\State Initialize $n$ indices $\{i|p(i)\sim uniform(n), i\notin \{0,n\}\}$
\\// \textcolor{gray}{Parallel the following process on $n$ events}
\\// \textcolor{gray}{For $i^{th}$ event:}
\For {$epoch=1 \rightarrow T_1^{max}$}
    \State $b=e=i$
    \For {$j=b \rightarrow 0$} //  \textcolor{gray}{or $j=e \rightarrow N-1$}
        \State $s_j=sim(F_j, [F_b,...,F_e])=cos(\mathbf{h}^F_j, \frac{1}{b-e+1}\sum_{k=b}^e \mathbf{h}^F_k)$
        \If {$s_j<\delta$}
            \State Break
        \Else
            \State $j=b-1$ //  \textcolor{gray}{or $j=e+1$}
        \EndIf
    \EndFor
\EndFor
\State Return $n$ events $E_1^1,E_2^1,...,E_n^1$
\end{algorithmic}

\noindent\rule{0.46\textwidth}{0.4pt}
\begin{algorithmic}[2]
\Require A video $V$, $n$ initial events $E_1^1,E_2^1,...,E_n^1$
\Ensure $n$ refined video events $(E_1^2,E_2^2,..,E_n^2)$
\\// \textcolor{gray}{Parallel the following process on $n$ events}
\For {$epoch=1 \rightarrow T_2^{max}$}
    \State Sample $m_v$ frames ${F_q|p(q)\sim uniform(N),\{b,e\}\in q}$
    \State Get each hidden representation $\mathbf{h}_q,=ViT(F_q)$
    \State Extract $m_l$ tuples $\{(a_1,p,a_2)_u\}^{m_l}$ based on RobustOIE
    \State Get hidden each representation $\mathbf{h}_u=BERT(concat((a_1,p,a_2)_u))$
    \State Calculate similarity:
        \State $\xi_t = \sum_{(u,q)\in\psi} sim(l_u, e_q)$
        \State $sim(l_u, e_q) = cos(\mathbf{h}_u, \mathbf{h}_q)$
    \If {$\xi_t < \xi_{t-1}$}
        \State Break
    \Else
        \State $b-=5$, $e+=5$
    \EndIf
\EndFor
\State Return $n$ events $E2_1^2,E_2^2,...,E_n^2$
\end{algorithmic}
\end{algorithm}
\end{small}

\section{Hyper-parameters Settings}

    \begin{table}[hpt]
 \setlength{\tabcolsep}{8pt}
    \centering
    \begin{tabular}{lr}
      \toprule[1.5pt]
        \textbf{Hyper-parameter} & \textbf{Value} \\
      \hline
        Confidence Threshold $\tau$ &  0.4 \\
        Vision Encoder in InsOVER-S1 & ViT-L/14 from CLIP \\
        $\delta_1$ in InsOVER-S1 & 0.95 \\
        FPS in InsOVER-S1 & 1 \\
        Language Encoder in InsOVER-S2 & BERT-base \\
        Vision Encoder in InsOVER-S2 & ViT-L/14 from CLIP \\
        FPS in InsOVER-S2 & 1 \\
        OpenIE Model in InsOVER-S2 & RobustOIE \\
        Number of sampling frames $m_v$ & 3,4,5,6 \\
      \hline
    \end{tabular}
    \caption{Settings of InsOVER algorithm}
    \label{tab:setting_base_network}
\end{table}

\section{Details of Prompts Design}\label{app:prompt}
\label{app:sec:details_of_prompts_design}

\begin{figure}[hpt]
    \centering
    \includegraphics[scale=0.55]{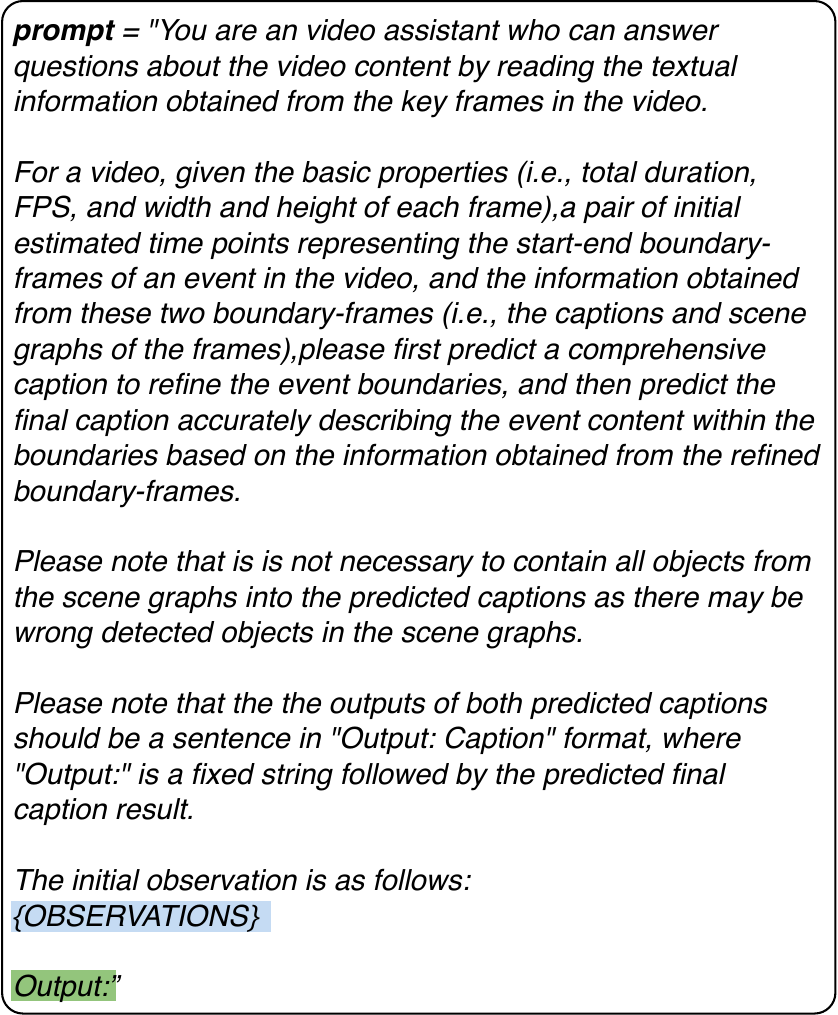}
    \caption{The prepend prompt for video comprehension with LLMs.}
    \label{fig:pre_prompt}
\end{figure}

\begin{figure}[hpt]
    \centering
    \includegraphics[scale=0.55]{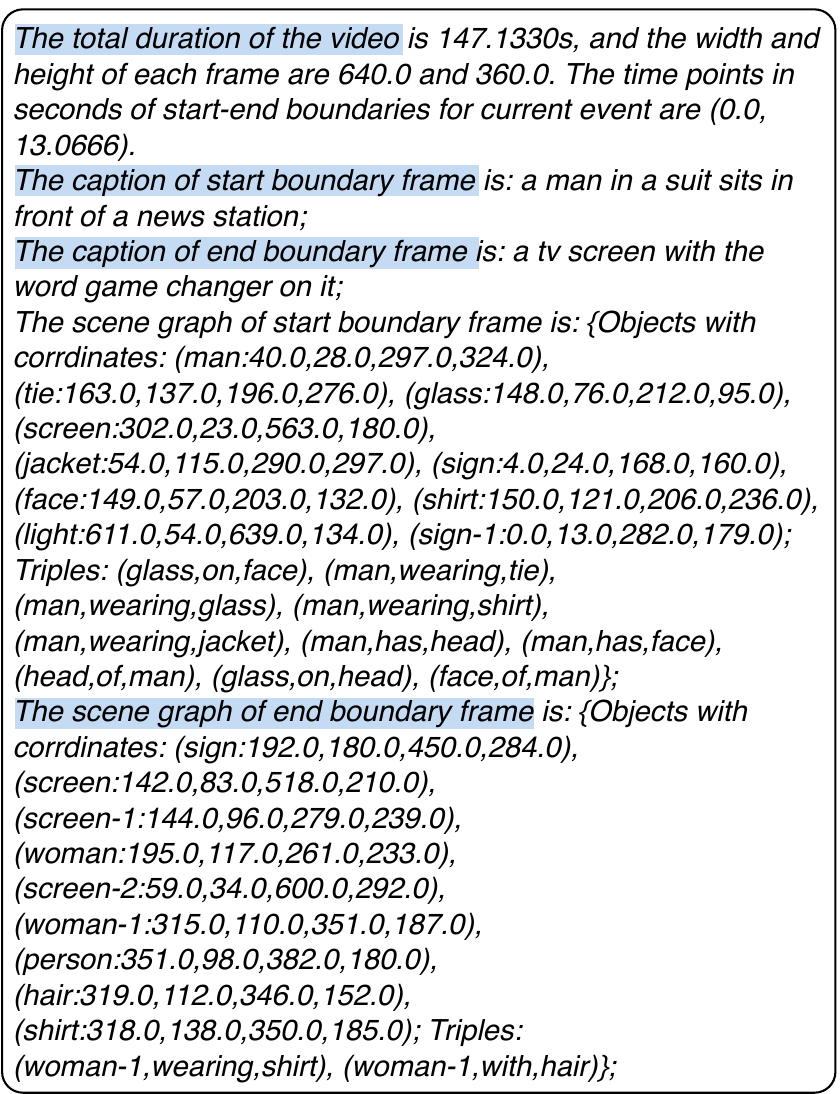}
    \caption{An example showing initial observations based on two visual tools.}
    \label{fig:enter-label}
\end{figure}


\end{document}